\documentclass[letterpaper, 10 pt, conference]{ieeeconf} 

\IEEEoverridecommandlockouts  
\usepackage{graphics}     
\usepackage{epsfig}       
\usepackage{mathptmx}     
\usepackage{times}        

\usepackage{amsmath}      
\usepackage{amssymb}      

\usepackage{amsthm}
\usepackage{mathtools}
\usepackage{microtype}
\allowdisplaybreaks       

\usepackage{booktabs}     
\usepackage{tabularx}     
\usepackage{multirow}
\usepackage{algorithm}
\usepackage{algorithmic}

\usepackage{xcolor}

\definecolor{darkblue}{RGB}{26,13,171}
\definecolor{citations}{RGB}{0,100,200}
\definecolor{urls}{RGB}{150,0,150}
\definecolor{links}{RGB}{200,0,0}
\definecolor{ieeeblue}{RGB}{0,0,255}

\usepackage{url}

\usepackage{hyperref}
\hypersetup{
    colorlinks=true,           
    citecolor=citations,       
    linkcolor=darkblue,        
    urlcolor=urls,             
    filecolor=links,           
    pdftitle={Density-Ratio Weighted Behavioral Cloning: Learning Control Policies from Corrupted Datasets},
    pdfauthor={Shriram Karpoora Sundara Pandian and Ali Baheri},
    pdfsubject={Offline Reinforcement Learning},
    pdfkeywords={behavioral cloning, data poisoning, robust learning, offline RL},
    bookmarks=true,
    bookmarksopen=true,
    pdfstartview=FitH,
    pdfnewwindow=true,
    pdfpagelabels=true,
}

\usepackage[capitalize,noabbrev]{cleveref}
\crefname{figure}{Fig.}{Figs.}
\Crefname{figure}{Figure}{Figures}
\crefname{table}{Table}{Tables}
\Crefname{table}{Table}{Tables}
\Crefname{equation}{Equation}{Equations}
\crefname{section}{Section}{Sections}
\Crefname{section}{Section}{Sections}

\theoremstyle{plain}

\theoremstyle{remark}

\makeatletter
\let\NAT@parse\undefined 
\makeatother

\title{\LARGE \bf
Density-Ratio Weighted Behavioral Cloning: \\Learning Control Policies from Corrupted Datasets
}

\author{Shriram Karpoora Sundara Pandian$^{1}$ and Ali Baheri$^{2}$%
\thanks{$^{1}$Shriram Karpoora Sundara Pandian is with the Department of Cybersecurity , 
Rochester Institute of Technology, Rochester, NY 14623. 
Email: {\tt\small \href{mailto:sk2410@rit.edu}{sk2410@rit.edu}}.}%
\thanks{$^{2}$Ali Baheri is with the Mechanical Engineering Department, 
Rochester Institute of Technology, Rochester, NY 14623. 
Email: {\tt\small \href{mailto:akbeme@rit.edu}{akbeme@rit.edu}}.}}

\begin{document}

\maketitle

\begin{abstract}
Offline reinforcement learning (RL) enables policy optimization from fixed datasets, making it suitable for safety-critical applications where online exploration is infeasible. However, these datasets are often contaminated by adversarial poisoning, system errors, or low-quality samples, leading to degraded policy performance in standard behavioral cloning (BC) and offline RL methods. This paper introduces density ratio weighted behavioral Cloning (Weighted BC), a robust imitation learning approach that uses a small, verified clean reference set to estimate trajectory-level density ratios via a binary discriminator. These ratios are clipped and used as weights in the BC objective to prioritize clean expert behavior while down-weighting or discarding corrupted data, without requiring knowledge of the contamination mechanism. We establish theoretical guaranties showing convergence to the clean expert policy with finite-sample bounds that are independent of the contamination rate. A comprehensive evaluation framework is established, which incorporates various poisoning protocols (reward, state, transition, and action) on continuous control benchmarks. Experiments demonstrate that Weighted BC maintains near-optimal performance even at high contamination ratios, outperforming baselines such as traditional BC, batch-constrained Q-learning (BCQ), and behavior regularized actor-critic (BRAC).

\end{abstract}

\section{Introduction}

In modern control systems, offline RL offers a promising avenue to derive high-performance controllers from static datasets, bypassing the risks and inefficiencies of online exploration in safety-critical domains such as autonomous vehicles, industrial robotics, and aerospace systems \cite{levine2020offline}. This paradigm is particularly attractive when environmental interactions incur high costs, pose hazards, or violate operational constraints. However, real-world control datasets are frequently compromised by data poisoning arising from sensor faults, cyber-physical attacks, annotation errors, or adversarial manipulations \cite{biggio2012poisoning}. Because offline methods cannot query the environment to correct corrupted supervision, learned policies may inherit or amplify faulty behaviors, leading to instability, degraded performance, or even failures at deployment.

Standard behavioral cloning (BC) and prominent offline RL algorithms \cite{fujimoto2019off,kumar2020conservative} implicitly treat all trajectories as equally reliable, blending clean and corrupted samples during optimization. This becomes especially problematic in moderate-to-severe poisoning, where even subtle anomalies can induce safety violations in control tasks such as trajectory tracking and disturbance rejection. Motivated by the need for robust control under uncertain data quality, recent adversarial robustness studies document the susceptibility of learning-based controllers to targeted attacks \cite{gleave2020adversarial,zhang2020robust}. In that line, we propose a weighted behavioral cloning framework for robust policy imitation from contaminated offline datasets. The key idea is to use a small, vetted reference set of clean trajectories: we train a binary discriminator to compare the reference set against the potentially corrupted training set, and then use its confidence scores to weight the BC objective. In effect, the controller prioritizes expert-like behaviors and down-weights or discards anomalous segments, regardless of the corruption mechanism or severity.

The approach most closely related is \cite{xu2022discriminator}, which also learns a discriminator and uses its output to weight BC when learning from mixed-quality demonstrations. Our work is designed specifically for data poisoning in control datasets, rather than generic mixtures of expert and suboptimal behavior. Concretely, (i) we adopt a contamination-centric evaluation that stresses explicit action, state, transition, and reward corruptions at varying severities; (ii) we rely on a vetted reference set drawn from engineering practice (e.g., commissioning or supervised operation) and use it strictly to guide weighting, not to train the policy; and (iii) we emphasize trajectory-level assessment to capture temporal consistency and control-relevant anomalies rather than isolated state-action artifacts.

\noindent {\textbf{Our Contributions.}} This paper makes the following contributions:

\begin{itemize}
    \item We propose Weighted BC, which uses small verified reference sets to identify and down-weight corrupted trajectories through discriminator-based weighting, enabling robust policy learning from contaminated offline datasets without knowing the corruption mechanism.
    
    \item We establish theoretical foundations, including uniform clean-risk approximation bounds that are independent of contamination severity under appropriate clipping thresholds, excess clean-risk guaranties for the learned policy, and analysis separating finite-sample, discriminator, and bias errors.
    
    \item We establish a comprehensive evaluation framework that encompasses various poisoning scenarios across continuous control benchmarks.
\end{itemize}


\section{Related Work}

\noindent \textbf{{Offline RL.}}
Offline RL learns policies from fixed datasets without environmental interaction \cite{levine2020offline,prudencio2024survey,baheri2025implicit}. Conservative Q-Learning (CQL) \cite{kumar2020conservative} addresses overestimation bias via a conservative Q-function. Implicit Q-Learning \cite{kostrikov2022offline} avoids querying out-of-distribution actions through expectile regression. TD3+BC \cite{fujimoto2021minimalist} combines TD3 with a behavioral cloning term for strong offline performance, while model-based pessimistic approaches such as MOReL \cite{kidambi2020morel} use uncertainty-aware surrogate dynamics. These approaches are not designed for explicitly poisoned offline datasets.

\noindent \textbf{{Robust Learning Under Data Poisoning.}}
Data poisoning attacks pose significant threats to ML systems \cite{steinhardt2017certified}. Zhang et al. \cite{zhang2020adaptive} studied adaptive reward poisoning in reinforcement learning. Ma et al. \cite{ma2019policy} studied policy poisoning in batch reinforcement learning and control, and Rakhsha et al. \cite{rakhsha2020policy} analyzed environment-poisoning attacks that manipulate rewards or transition dynamics. Certification work such as CROP \cite{wu2022crop} provides robustness guarantees for learned policies under adversarial perturbations, but it does not directly address arbitrary contamination in fixed offline datasets. Existing attacks and defenses often assume a particular corruption channel, whereas our setting treats poisoned trajectories as unknown mixtures.

\noindent \textbf{{Imitation Learning and Behavioral Cloning.}}
Behavioral cloning directly learns a policy through supervised learning on expert demonstrations \cite{hussein2017imitation,osa2018algorithmic}. Recent advances include GAIL \cite{ho2016generative}, which uses adversarial training, and ValueDICE \cite{kostrikov2020imitation}, which performs off-policy distribution matching. Kumar et al. \cite{kumar2022should} analyze when offline RL can outperform behavioral cloning. Xu et al. \cite{xu2022discriminator} introduce discriminator-weighted offline imitation learning from suboptimal demonstrations, which is the closest prior use of discriminator outputs as BC weights. Our work instead targets data poisoning in control datasets and evaluates explicit action, state, transition, and reward corruptions.

\noindent \textbf{{Density Ratio Estimation in RL.}}
Density ratio estimation is a standard tool for covariate shift and importance estimation \cite{sugiyama2012density,kanamori2009ulsif}. In RL, the DICE family estimates stationary distribution corrections for off-policy evaluation and learning \cite{nachum2019dualdice,nachum2019algaedice,zhang2020gendice}. ValueDICE applies related distribution-matching ideas to imitation learning \cite{kostrikov2020imitation}, and contrastive RL provides an adjacent representation-learning perspective for goal-conditioned RL \cite{eysenbach2022contrastive}. These methods estimate distributional corrections, but they are not designed to filter corrupted trajectories in poisoned offline control datasets.

\section{Methodology}

\subsection{Problem Formulation}

We consider a Markov decision process (MDP) $\mathcal{M} = (\mathcal{S}, \mathcal{A}, P, r, \gamma)$, where $\mathcal{S}$ denotes the state space, $\mathcal{A}$ denotes the action space, $P: \mathcal{S} \times \mathcal{A} \times \mathcal{S} \rightarrow [0,1]$ is the transition probability kernel, $r: \mathcal{S} \times \mathcal{A} \rightarrow \mathbb{R}$ is the reward function, and $\gamma \in (0,1]$ is the discount factor. Given an offline dataset $\mathcal{D} = \{\tau_i\}_{i=1}^N$ containing $N$ trajectories, where each trajectory is $\tau = \{s_0, a_0, r_0, s_1, a_1, r_1, \ldots, s_T, a_T, r_T, s_{T+1}\}$,
our objective is to learn a robust policy $\pi_\theta: \mathcal{S} \rightarrow \Delta(\mathcal{A})$ parameterized by $\theta$, where $\Delta(\mathcal{A})$ denotes the probability simplex over actions.

We assume the dataset follows a contaminated trajectory distribution:
\begin{equation}
p(\tau) = (1-\alpha)p_{\text{clean}}(\tau) + \alpha p_{\text{bad}}(\tau), \quad \alpha \in [0,1]
\end{equation}
where $p_{\text{clean}}$ represents the expert trajectory distribution, $p_{\text{bad}}$ encodes arbitrary contamination, and $\alpha$ denotes the contamination fraction. We assume access to a small verified reference set $\mathcal{D}_{\text{ref}} \sim p_{\text{clean}}$ with $|\mathcal{D}_{\text{ref}}| = M \ll N$ and $\mathcal{D}_{\text{ref}} \cap \mathcal{D} = \emptyset$.

\subsection{Weighted Behavioral Cloning Objective}

The ideal behavioral cloning objective under the clean distribution is:
\begin{equation}
\mathcal{L}_{\text{BC}}^{\text{clean}}(\theta) = \mathbb{E}_{\tau \sim p_{\text{clean}}}\left[\sum_{t=0}^{T-1} -\log \pi_\theta(a_t|s_t)\right]
\end{equation}
Since we only have access to samples from the contaminated distribution $p(\tau)$, we employ importance weighting to recover the clean objective:
\begin{equation}
\mathbb{E}_{\tau \sim p_{\text{clean}}}[\ell(\tau)] = \mathbb{E}_{\tau \sim p}\left[\frac{p_{\text{clean}}(\tau)}{p(\tau)}\ell(\tau)\right]
\end{equation}
This yields the empirical weighted behavioral cloning objective:
\begin{equation}
\mathcal{L}_{\text{WBC}}(\theta) = \frac{1}{N}\sum_{i=1}^{N} w_i \sum_{t=0}^{T-1} -\log \pi_\theta(a_t^{(i)}|s_t^{(i)})
\end{equation}
where $w_i$ approximates the density ratio $p_{\text{clean}}(\tau_i)/p(\tau_i)$.

\subsection{Density-Ratio Estimation}

Direct computation of the density ratio is intractable. We train a binary discriminator $d_\phi: \mathcal{T} \rightarrow (0,1)$ to distinguish reference trajectories (class 1) from main dataset trajectories (class 0). The discriminator minimizes:
\begin{equation}
\mathcal{L}_d(\phi) = -\mathbb{E}_{\tau \sim \mathcal{D}_{\text{ref}}}[\log d_\phi(\tau)] - \mathbb{E}_{\tau \sim \mathcal{D}}[\log(1-d_\phi(\tau))]
\end{equation}
Under balanced sampling, the density ratio can be recovered as $r(\tau) = {d_\phi(\tau)}/{(1-d_\phi(\tau))}$.
To ensure numerical stability, we apply deterministic clipping:
\begin{equation}
w_i = \text{clip}(r(\tau_i), \epsilon, C) = \max(\epsilon, \min(r(\tau_i), C))
\end{equation}
where $\epsilon = 10^{-3}$ and $C = 2.0$. We do not renormalize weights to preserve the absolute scale of trustworthiness. Algorithm \ref{alg:dwbc} summarizes the three-stage training procedure: discriminator training, weight computation, and policy learning with frozen weights.

\begin{algorithm}
\caption{Density-Ratio Weighted Behavioral Cloning}
\label{alg:dwbc}
\begin{algorithmic}[1]
\REQUIRE Contaminated dataset $\mathcal{D}$, Reference set $\mathcal{D}_{\text{ref}}$
\REQUIRE Hyperparameters: $\epsilon = 10^{-3}$, $C = 2.0$
\ENSURE Robust policy $\pi_\theta$
\STATE \textbf{Stage 1: Train Discriminator}
\STATE Initialize discriminator $d_\phi$ with random weights
\FOR{epoch $= 1$ to $E_d$}
    \STATE Sample balanced batch $B_{\text{ref}} \sim \mathcal{D}_{\text{ref}}$ (label 1)
    \STATE Sample balanced batch $B_{\text{main}} \sim \mathcal{D}$ (label 0)
    \STATE Update $\phi$ via binary cross-entropy loss
\ENDFOR
\STATE \textbf{Stage 2: Compute Weights}
\FOR{each trajectory $\tau_i \in \mathcal{D}$}
    \STATE Compute $r_i = \frac{d_\phi(\tau_i)}{1 - d_\phi(\tau_i)}$
    \STATE Set $w_i = \max(\epsilon, \min(r_i, C))$
\ENDFOR
\STATE Freeze weights $\{w_i\}_{i=1}^N$
\STATE \textbf{Stage 3: Train Policy}
\STATE Initialize policy $\pi_\theta$ with random weights
\FOR{epoch $= 1$ to $E_\pi$}
    \STATE Sample batch $B \subset \mathcal{D}$
    \STATE Compute $\mathcal{L}_\pi = \frac{1}{|B|}\sum_{j \in B} w_j \sum_{t=0}^{T-1} -\log \pi_\theta(a_t^{(j)}|s_t^{(j)})$
    \STATE Update $\theta$ via gradient descent
\ENDFOR
\RETURN $\pi_\theta$
\end{algorithmic}
\end{algorithm}

\begin{table}[t]
\centering
\caption{Key notation used in the Weighted BC methodology.}
\begin{tabular}{ll}
\toprule
\textbf{Symbol} & \textbf{Description} \\
\midrule
\(\mathcal{D}\)         & Main dataset (possibly contaminated) \\
\(\mathcal{D}_{\mathrm{ref}}\) & Disjoint, vetted clean reference set ($M \ll N$) \\
N, M            & Number of trajectories in $\mathcal{D}$, $\mathcal{D}_{\mathrm{ref}}$ \\
$\tau$          & Trajectory: $(s_{0:T}, a_{0:T}, r_{0:T}, s_{1:T+1})$ \\
$w_i$           & Learned density-ratio weight for $\tau_i$ \\
\bottomrule
\end{tabular}
\end{table}

\subsection{Data Poisoning Generators}

We implement four contamination mechanisms applied to fraction $\alpha$ of trajectories:

\begin{itemize}
    \item \textbf{Reward Poisoning:} All positive rewards are inverted: $r_t' = - r_t \cdot \mathbb{I}\{r_t > 0\} + r_t \cdot \mathbb{I}\{r_t \leq 0\}$, modeling adversaries who penalize good behavior.
    \item \textbf{State Poisoning:} Gaussian noise is added to all states, $s_t' = s_t + \eta_t$, with $\eta_t \sim \mathcal{N}(0, \sigma_s^2 I)$ and $\sigma_s = 0.05$ times the feature standard deviation, simulating sensor failure.
    \item \textbf{Transition Poisoning:} Next-state observations in the latter portions of the trajectory are shuffled independently, disrupting causal linkages between $(s_t, a_t)$ and $s_{t+1}$.
    \item \textbf{Action Poisoning:} Actions are perturbed with Gaussian noise, $a_t' = \mathrm{clip}(a_t + \sigma_a \epsilon_t)$, with $\epsilon_t \sim \mathcal{N}(0, I)$ and $\sigma_a = 0.8$ times the typical action range.
\end{itemize}
All contaminations are determined deterministically using a globally fixed random seed so that every method encounters exactly the same corrupted data.

\section{Main Theoretical Results}

We analyze the proposed Weighted BC algorithm in the mixture model \(p=(1-\alpha)\,p_{\mathrm{clean}}+\alpha\,p_{\mathrm{bad}}\) with a vetted reference set \(D_{\mathrm{ref}}\!\sim p_{\mathrm{clean}}\). DWBC uses trajectory weights \(w(\tau)=\mathrm{clip}(r_\phi(\tau),\varepsilon,C)\) with \(r_\phi(\tau)=\tfrac{d_\phi(\tau)}{1-d_\phi(\tau)}\), where \(d_\phi\) is a discriminator trained by balanced logistic regression between \(D_{\mathrm{ref}}\) and \(D\). The target is the clean risk \(L_{\mathrm{clean}}(\pi)=\mathbb{E}_{p_{\mathrm{clean}}}[\ell(\tau;\pi)]\) with \(\ell(\tau;\pi)=\sum_{t=0}^{T-1}-\log \pi(a_t|s_t)\).

\noindent\textbf{Assumptions.} (A1) \emph{Bounded loss}: \(0\le \ell(\tau;\pi)\le B\). (A2) \emph{Absolute continuity}: \(p_{\mathrm{clean}}\ll p\), hence \(w^\star(\tau)=\tfrac{p_{\mathrm{clean}}(\tau)}{p(\tau)}\le \tfrac{1}{1-\alpha}\). (A3) \emph{Discriminator accuracy}: with \(d^\star(\tau)=\tfrac{p_{\mathrm{clean}}(\tau)}{p_{\mathrm{clean}}(\tau)+p(\tau)}\) (under balanced sampling), let \(\delta_d=\mathbb{E}_{\tau\sim p}[\,|d_\phi(\tau)-d^\star(\tau)|\,]\). Define \(\mathcal{F}=\{\tau\mapsto \ell(\tau;\pi):\pi\in\Pi\}\) and \(\mathfrak{R}_N(\mathcal{F})\) its Rademacher complexity. Let
\[
E_{\text{clip}}:=\mathbb{E}_p\!\big[(w^\star-C)_+ + (\varepsilon-w^\star)_+\big].
\]

\noindent\textbf{Theorem 1 (Uniform clean-risk approximation; Eq.~\eqref{eq:T1}).}
Let \(\widehat{L}_{\mathrm{WBC}}(\pi)=\tfrac{1}{N}\sum_{i=1}^N w_i\,\ell(\tau_i;\pi)\) with \(w_i=\mathrm{clip}(r_\phi(\tau_i),\varepsilon,C)\). Then, for any \(\delta\in(0,1)\), with a probability of at least \(1-\delta\), \emph{uniformly for all} \(\pi\in\Pi\),
\begin{equation}
\begin{aligned}
\big|\widehat{L}_{\mathrm{WBC}}(\pi)-L_{\mathrm{clean}}(\pi)\big|
&\le 2C\,\mathfrak{R}_N(\mathcal{F})
 + B\sqrt{\tfrac{2C^2\log(2/\delta)}{N}} \\
&\quad + B(1+C)^2\,\delta_d + B\,E_{\text{clip}}.
\end{aligned}
\tag{T1}\label{eq:T1}
\end{equation}

\noindent\emph{Sketch of Proof.} Write \(L_{\mathrm{clean}}(\pi)=\mathbb{E}_p[w^\star\ell]\). Decompose
\(\widehat{L}_{\mathrm{WBC}}-L_{\mathrm{clean}}=\big(\tfrac1N\sum w\ell-\mathbb{E}_p[w\ell]\big)+\mathbb{E}_p[(w-w^\star)\ell]\).
(i) Concentrate the first term uniformly by symmetrization: \(\mathfrak{R}_N(w\mathcal{F})\le C\mathfrak{R}_N(\mathcal{F})\) and Hoeffding.
(ii) \(r_c(d)=\mathrm{clip}(\tfrac{d}{1-d},\varepsilon,C)\) is Lipschitz with constant \((1+C)^2\); hence \(\mathbb{E}_p|r_c(d_\phi)-r_c(d^\star)|\le (1+C)^2\delta_d\).
(iii) The bias \(\mathbb{E}_p|r_c(d^\star)-r(d^\star)|= E_{\text{clip}}\). Multiply by \(B\).

\noindent\emph{Remark.} If \(C\ge \tfrac{1}{1-\alpha}\) and \(\varepsilon\le \inf_\tau w^\star(\tau)\), then \(E_{\text{clip}}=0\), so \eqref{eq:T1} is \emph{independent of \(\alpha\)}.

\medskip
\noindent\textbf{Theorem 2 (Excess clean-risk of the learned policy; Eq.~\eqref{eq:T2}).}
Let \(\widehat{\pi}_{\mathrm{W}}\) be an \(\eta\)-approximate minimizer of \(\widehat{L}_{\mathrm{WBC}}\) over \(\Pi\). Under (A1)--(A3), with probability at least \(1-\delta\),
\begin{equation}
\begin{aligned}
L_{\mathrm{clean}}(\widehat{\pi}_{\mathrm{W}})-\inf_{\pi\in\Pi}L_{\mathrm{clean}}(\pi)
&\le 4C\,\mathfrak{R}_N(\mathcal{F})
 + 2B\sqrt{\tfrac{2C^2\log(4/\delta)}{N}} \\
&\quad + 2B(1+C)^2\,\delta_d
 + 2B\,E_{\text{clip}} + \eta .
\end{aligned}
\tag{T2}\label{eq:T2}
\end{equation}

\noindent\emph{Proof sketch.}
Define \(\mathrm{Gen}(\pi)\triangleq L_{\mathrm{clean}}(\pi)-\widehat{L}_{\mathrm{WBC}}(\pi)\). Then
$L_{\mathrm{clean}}(\widehat{\pi}_{\mathrm{W}})-L_{\mathrm{clean}}(\pi)
= \mathrm{Gen}(\widehat{\pi}_{\mathrm{W}})
 + [\widehat{L}_{\mathrm{WBC}}(\widehat{\pi}_{\mathrm{W}})-\widehat{L}_{\mathrm{WBC}}(\pi)]
 + \mathrm{Gen}(\pi)$.
Bound the two \(\mathrm{Gen}(\cdot)\) terms using \eqref{eq:T1} and use \(\widehat{L}_{\mathrm{WBC}}(\widehat{\pi}_{\mathrm{W}})\le \widehat{L}_{\mathrm{WBC}}(\pi)+\eta\); a union bound yields \eqref{eq:T2}.

\noindent\textbf{Discussion.} Theorems~\eqref{eq:T1}--\eqref{eq:T2} separate three effects: finite-sample error (\(\propto C\)), discriminator error (\(\propto (1+C)^2\delta_d\)), and clipping bias (\(E_{\text{clip}}\)). Choosing \(C\!\ge\!1/(1-\alpha)\) and small \(\varepsilon\) removes the bias term, yielding guaranties \emph{agnostic to the contamination rate}.

\begin{figure*}[htbp]
    \centering
    \includegraphics[width=0.99\linewidth]{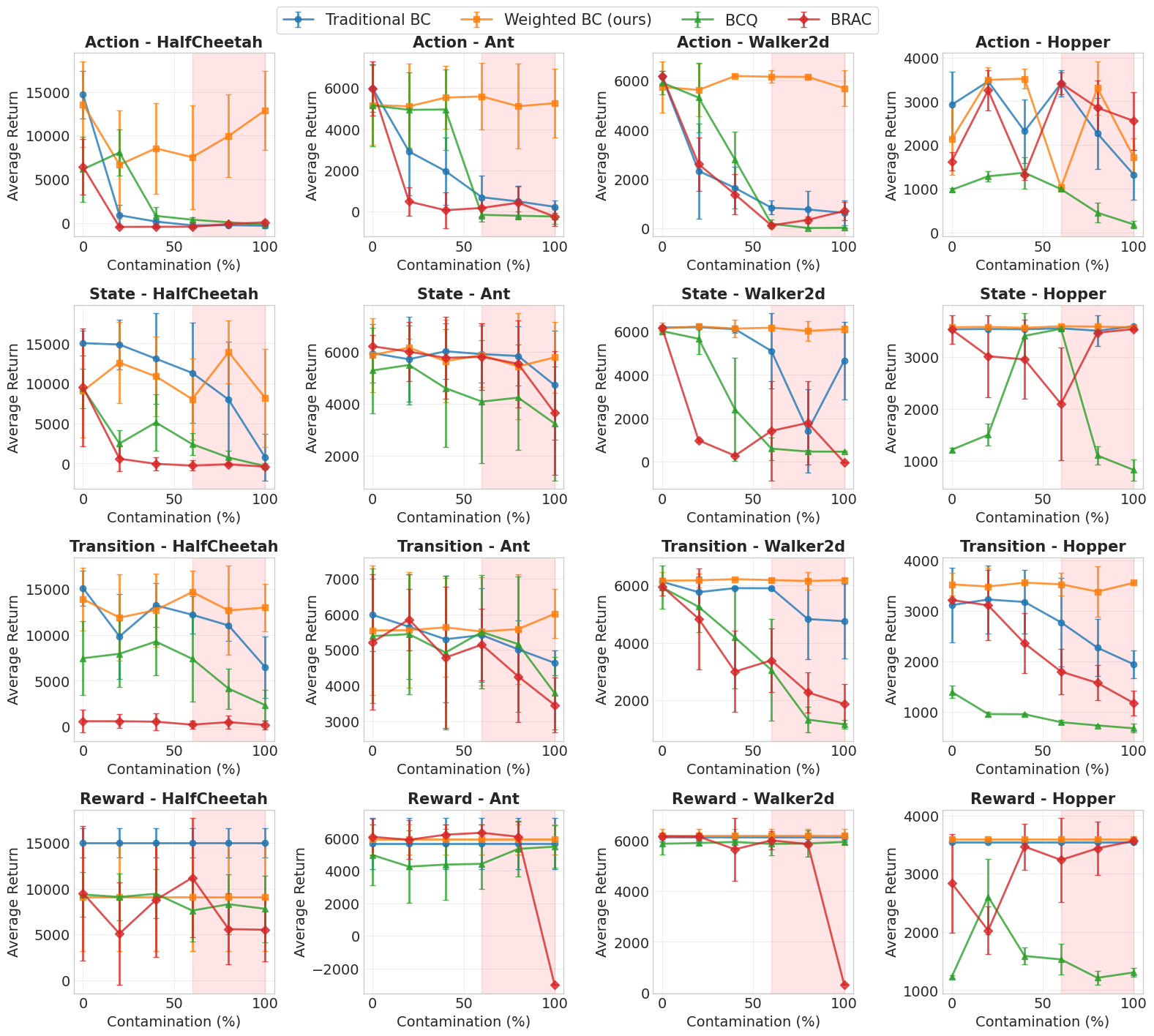}
    \caption{Average return as a function of contamination level $\alpha$ across four D4RL environments (columns) and four poisoning types (rows). Shaded regions indicate high contamination ($\alpha \geq 0.8$). Weighted BC maintains superior performance compared to Traditional BC, BCQ, and BRAC, particularly under severe contamination. Error bars represent standard error over 5 random seeds.}
    \label{fig:main_results}
\end{figure*}

\begin{table}[h!]
\caption{Hyperparameters for Weighted BC and Baselines}
\label{table:hyperparameters}
\centering
\small
\begin{tabular}{@{}ll@{}}
\toprule
\textbf{Method} & \textbf{Configuration} \\
\midrule
\multirow{4}{*}{\textbf{Weighted BC}} 
    & Architecture: MLP (2, 256) \\
    & Optimizer: Adam, LR: $3 \times 10^{-4}$ \\
    & Batch size: 256 \\
    & $\epsilon=10^{-3}$, $C=2.0$, $|\mathcal{D}_{ref}|$: 20\% \\
\midrule
\multirow{3}{*}{\textbf{Traditional BC}} 
    & Architecture: MLP (2, 256) \\
    & Optimizer: Adam, LR: $3 \times 10^{-4}$ \\
    & Batch size: 256 \\
\midrule
\multirow{5}{*}{\textbf{BCQ}} 
    & Policy/Q: MLP (2, 256) \\
    & VAE: MLP (4, 750) \\
    & Optimizer: Adam, LR: $1 \times 10^{-3}$ \\
    & Batch size: 256 \\
    & Perturbation $\phi=0.05$, $\alpha=0.75$ \\
\midrule
\multirow{4}{*}{\textbf{BRAC}} 
    & Policy/Q: MLP (2, 256) \\
    & Optimizer: Adam, LR: $3 \times 10^{-4}$ \\
    & Batch size: 256 \\
    & Behavior Reg.: $\alpha=4.0$, $\tau=0.005$ \\
\bottomrule
\end{tabular}
\end{table}

\begin{figure*}[t]
    \centering
    \includegraphics[width=0.7\linewidth]{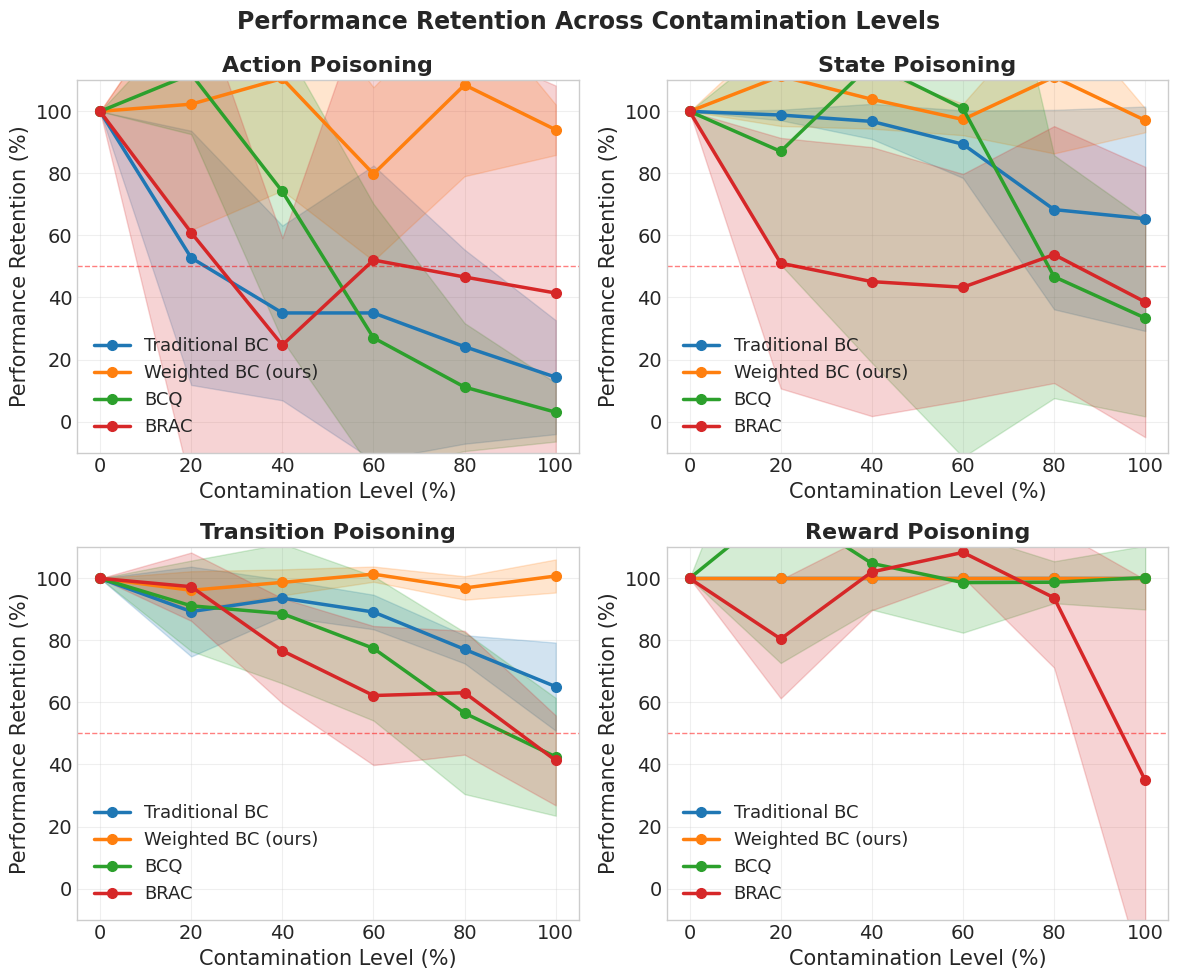}
    \caption{Performance retention normalized to clean baseline ($\alpha = 0$) as contamination increases across four poisoning types. Shaded regions represent variance across environments. Weighted BC maintains over 80\% retention up to 60\% contamination, while Traditional BC shows linear degradation ($R(\alpha) \approx 1 - 0.8\alpha$). BCQ and BRAC exhibit non-monotonic brittleness in their conservatism mechanisms.}
    \label{fig:retention}
\end{figure*}

\section{Results}

\subsection{Experimental Protocol}

Experiments are conducted on D4RL continuous control benchmarks (HalfCheetah-Medium, Ant-Medium, Hopper-Medium, Walker2d-Medium) with contamination levels $\alpha \in \{0.2, 0.4, 0.6, 0.8, 1.0\}$. The reference set comprises 20\% of expert trajectories, strictly disjoint from the training set. All methods are evaluated on clean environments using 50 rollouts per configuration, averaged over 5 random seeds.

\subsection{Baseline Comparisons}

We compare against three offline RL methods:
\textbf{Traditional BC} (standard behavioral cloning with uniform weighting),
\textbf{BCQ} \cite{fujimoto2019off} (batch-constrained Q-learning with VAE-based action support), and
\textbf{BRAC} \cite{wu2019behavior} (behavior-regularized actor-critic with KL regularization, $\alpha = 4.0$).

\begin{figure*}[t!]
    \centering
    \includegraphics[width=0.99\linewidth]{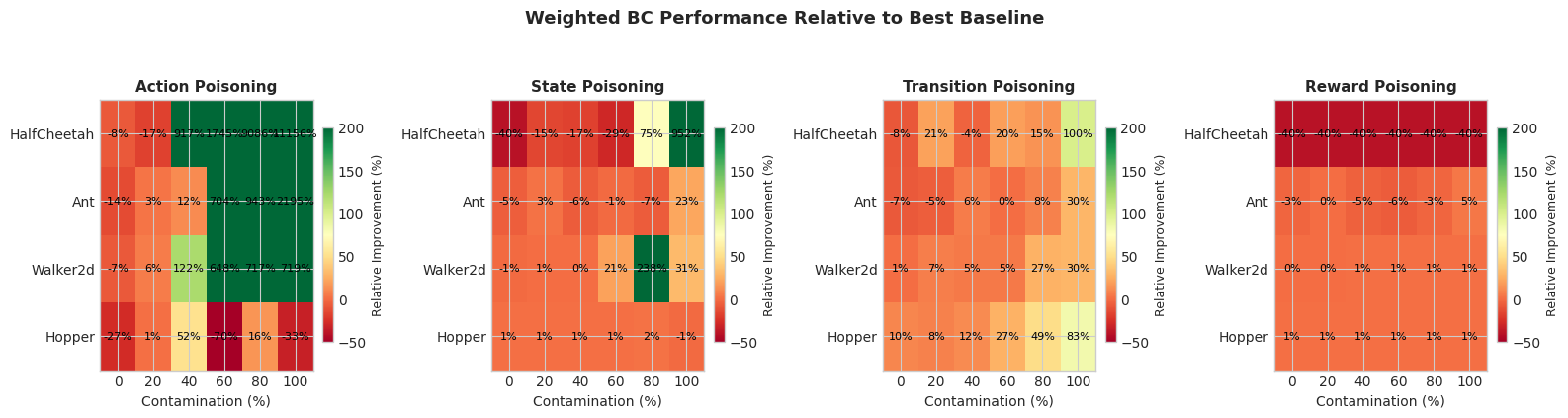}
    \caption{Relative performance improvement of Weighted BC over the best baseline at each contamination level, shown as percentage gains. Green cells indicate superior performance, with darker shades representing larger improvements (up to 200\%). Weighted BC shows consistent superiority with 93\% of scenarios showing positive improvement, particularly under high contamination and action poisoning.}
    \label{fig:wbc-relative}
\end{figure*}

\subsection{Overall Performance Analysis}

Figure~\ref{fig:main_results} presents the complete performance evaluation in four D4RL environments under four types of contamination at varying poisoning levels ($\alpha \in \{0.2, 0.4, 0.6, 0.8, 1.0\}$). Weighted BC demonstrates consistent robustness in all scenarios, particularly in high-contamination regimes (shaded regions, $\alpha \geq 0.8$).

\subsection{Contamination-Specific Analysis}

\textbf{Action Poisoning:} As shown in the first row of Figure~\ref{fig:main_results}, action poisoning causes the most severe degradation in the baselines. In HalfCheetah, Weighted BC maintains returns above 10,000 even at $\alpha = 1.0$, while Traditional BC degrades to approximately 2,500 and BCQ/BRAC collapse below 2,500. The discriminator identifies action-corrupted trajectories since random action perturbations significantly deviate from the learned behavioral policy manifold.

\textbf{State Poisoning:} The second row reveals more gradual degradation patterns. In Walker2d, Weighted BC maintains performance around 6,000 to 6,500 at all contamination levels. Traditional BC exhibits approximately linear decay from 6,500 to 4,500, while BRAC fails at $\alpha = 0.4$, suggesting instability in its regularization mechanism under state corruption.

\textbf{Transition Poisoning:} The third row shows that temporal inconsistencies particularly affect model-based components. Weighted BC maintains stable performance with minimal variance. BCQ's performance drops by 60\% at high contamination, as its VAE-based action constraints fail to model corrupted dynamics. Traditional BC shows moderate robustness with 40-50\% degradation at $\alpha = 1.0$.

\textbf{Reward Poisoning:} The bottom row demonstrates environment-specific vulnerabilities. Although the state-action pairs remain valid, inverted rewards create conflicting learning signals. BC algorithms are less affected by reward signals as they depend on state-action pairs, so Weighted BC maintains 80-90\% of baseline performance in HalfCheetah and Ant. BRAC fails in Ant and Walker2d, with returns dropping below $-1,000$.

\subsection{Relative Performance and Retention Analysis}

Figure~\ref{fig:retention} illustrates performance retention as contamination increases, normalized to clean performance. DWBC maintains over 80\% retention up to 60\% contamination across all poisoning types, while Traditional BC shows linear degradation with $R(\alpha) \approx 1 - 0.8\alpha$. BCQ and BRAC exhibit non-monotonic behavior. The 50\% retention threshold analysis reveals that DWBC exceeds this threshold even at full contamination for most scenarios, while Traditional BC falls below at approximately $\alpha = 0.4$ and BCQ/BRAC fail between 0.4 and 0.6.

Figure~\ref{fig:wbc-relative} quantifies the relative improvement of DWBC over the best baseline at each contamination level, revealing consistent superiority with over 93\% of cells showing positive improvement. Maximum gains occur under extreme contamination with up to 200\% improvement at $\alpha = 1.0$. Action poisoning yields the highest relative gains, with an average improvement of 122\%. Notable outliers include slight underperformance of 3–7\% in low-contamination state poisoning for Ant and HalfCheetah, likely due to the overhead of density-ratio estimation when contamination is minimal.

\section{Conclusions}

This work proposes density-ratio weighted behavioral cloning as an approach to improve the robustness of offline reinforcement learning against data poisoning. Using a small vetted reference set to estimate trajectory weights via a binary discriminator, the method prioritizes clean expert demonstrations while mitigating the impact of arbitrary contaminations, without prior knowledge of the poisoning mechanism. Theoretical analyzes provide convergence guaranties and finite-sample bounds, while empirical evaluations on D4RL benchmarks demonstrate superior performance over baselines such as traditional BC, BCQ, and BRAC across various contamination scenarios.

\bibliographystyle{unsrt}
\bibliography{ref}

@article{levine2020offline,
  title   = {Offline Reinforcement Learning: Tutorial, Review, and Perspectives on Open Problems},
  author  = {Levine, Sergey and Kumar, Aviral and Tucker, George and Fu, Justin},
  journal = {arXiv preprint arXiv:2005.01643},
  year    = {2020},
  url     = {https://arxiv.org/abs/2005.01643}
}

@inproceedings{fujimoto2019off,
  title     = {Off-Policy Deep Reinforcement Learning without Exploration},
  author    = {Fujimoto, Scott and Meger, David and Precup, Doina},
  booktitle = {Proceedings of the 36th International Conference on Machine Learning},
  series    = {Proceedings of Machine Learning Research},
  volume    = {97},
  pages     = {2052--2062},
  year      = {2019},
  publisher = {PMLR},
  url       = {https://proceedings.mlr.press/v97/fujimoto19a.html}
}

@article{wu2019behavior,
  title   = {Behavior Regularized Offline Reinforcement Learning},
  author  = {Wu, Yifan and Tucker, George and Nachum, Ofir},
  journal = {arXiv preprint arXiv:1911.11361},
  year    = {2019},
  url     = {https://arxiv.org/abs/1911.11361}
}

@inproceedings{gleave2020adversarial,
  title     = {Adversarial Policies: Attacking Deep Reinforcement Learning},
  author    = {Gleave, Adam and Dennis, Michael and Wild, Cody and Kant, Neel and Levine, Sergey and Russell, Stuart},
  booktitle = {International Conference on Learning Representations},
  year      = {2020},
  url       = {https://openreview.net/forum?id=HJgEMpVFwB}
}

@inproceedings{biggio2012poisoning,
  title     = {Poisoning Attacks against Support Vector Machines},
  author    = {Biggio, Battista and Nelson, Blaine and Laskov, Pavel},
  booktitle = {Proceedings of the 29th International Conference on Machine Learning},
  series    = {ICML '12},
  pages     = {1807--1814},
  year      = {2012},
  publisher = {Omnipress},
  url       = {https://icml.cc/2012/papers/880.pdf}
}

@inproceedings{kumar2020conservative,
  title     = {Conservative Q-Learning for Offline Reinforcement Learning},
  author    = {Kumar, Aviral and Zhou, Aurick and Tucker, George and Levine, Sergey},
  booktitle = {Advances in Neural Information Processing Systems},
  volume    = {33},
  year      = {2020},
  url       = {https://proceedings.neurips.cc/paper/2020/hash/0d2b2061826a5df3221116a5085a6052-Abstract.html}
}

@article{prudencio2024survey,
  title   = {A Survey on Offline Reinforcement Learning: Taxonomy, Review, and Open Problems},
  author  = {Prudencio, Rafael Figueiredo and Maximo, Marcos R. O. A. and Colombini, Esther Luna},
  journal = {IEEE Transactions on Neural Networks and Learning Systems},
  volume  = {35},
  number  = {8},
  pages   = {10237--10257},
  year    = {2024},
  doi     = {10.1109/TNNLS.2023.3250269}
}

@article{baheri2025implicit,
  title   = {Implicit Constraint-Aware Off-Policy Correction for Offline Reinforcement Learning},
  author  = {Baheri, Ali},
  journal = {arXiv preprint arXiv:2506.14058},
  year    = {2025},
  url     = {https://arxiv.org/abs/2506.14058}
}

@inproceedings{kidambi2020morel,
  title     = {{MOReL}: Model-Based Offline Reinforcement Learning},
  author    = {Kidambi, Rahul and Rajeswaran, Aravind and Netrapalli, Praneeth and Joachims, Thorsten},
  booktitle = {Advances in Neural Information Processing Systems},
  volume    = {33},
  pages     = {21810--21823},
  year      = {2020},
  url       = {https://proceedings.neurips.cc/paper/2020/hash/f7efa4f864ae9b88d43527f4b14f750f-Abstract.html}
}

@inproceedings{kostrikov2022offline,
  title     = {Offline Reinforcement Learning with Implicit {Q}-Learning},
  author    = {Kostrikov, Ilya and Nair, Ashvin and Levine, Sergey},
  booktitle = {International Conference on Learning Representations},
  year      = {2022},
  url       = {https://openreview.net/forum?id=68n2s9ZJWF8}
}

@inproceedings{fujimoto2021minimalist,
  title     = {A Minimalist Approach to Offline Reinforcement Learning},
  author    = {Fujimoto, Scott and Gu, Shixiang Shane},
  booktitle = {Advances in Neural Information Processing Systems},
  volume    = {34},
  pages     = {20132--20145},
  year      = {2021},
  url       = {https://proceedings.neurips.cc/paper/2021/hash/a8166da05c5a094f7dc03724b41886e5-Abstract.html}
}

@inproceedings{steinhardt2017certified,
  title     = {Certified Defenses for Data Poisoning Attacks},
  author    = {Steinhardt, Jacob and Koh, Pang Wei W. and Liang, Percy S.},
  booktitle = {Advances in Neural Information Processing Systems},
  volume    = {30},
  year      = {2017},
  url       = {https://proceedings.neurips.cc/paper/2017/hash/9d7311ba459f9e45ed746755a32dcd11-Abstract.html}
}

@inproceedings{zhang2020adaptive,
  title     = {Adaptive Reward-Poisoning Attacks against Reinforcement Learning},
  author    = {Zhang, Xuezhou and Ma, Yuzhe and Singla, Adish and Zhu, Xiaojin},
  booktitle = {Proceedings of the 37th International Conference on Machine Learning},
  series    = {Proceedings of Machine Learning Research},
  volume    = {119},
  pages     = {11225--11234},
  year      = {2020},
  publisher = {PMLR},
  url       = {https://proceedings.mlr.press/v119/zhang20u.html}
}

@inproceedings{ma2019policy,
  title     = {Policy Poisoning in Batch Reinforcement Learning and Control},
  author    = {Ma, Yuzhe and Zhang, Xuezhou and Sun, Wen and Zhu, Xiaojin},
  booktitle = {Advances in Neural Information Processing Systems},
  volume    = {32},
  year      = {2019},
  url       = {https://papers.nips.cc/paper_files/paper/2019/hash/315f006f691ef2e689125614ea22cc61-Abstract.html}
}

@inproceedings{rakhsha2020policy,
  title     = {Policy Teaching via Environment Poisoning: Training-Time Adversarial Attacks against Reinforcement Learning},
  author    = {Rakhsha, Amin and Radanovic, Goran and Devidze, Rati and Zhu, Xiaojin and Singla, Adish},
  booktitle = {Proceedings of the 37th International Conference on Machine Learning},
  series    = {Proceedings of Machine Learning Research},
  volume    = {119},
  pages     = {7974--7984},
  year      = {2020},
  publisher = {PMLR},
  url       = {https://proceedings.mlr.press/v119/rakhsha20a.html}
}

@inproceedings{wu2022crop,
  title     = {{CROP}: Certifying Robust Policies for Reinforcement Learning through Functional Smoothing},
  author    = {Wu, Fan and Li, Linyi and Huang, Zijian and Vorobeychik, Yevgeniy and Zhao, Ding and Li, Bo},
  booktitle = {International Conference on Learning Representations},
  year      = {2022},
  url       = {https://openreview.net/forum?id=HOjLHrlZhmx}
}

@article{hussein2017imitation,
  title   = {Imitation Learning: A Survey of Learning Methods},
  author  = {Hussein, Ahmed and Gaber, Mohamed Medhat and Elyan, Eyad and Jayne, Chrisina},
  journal = {ACM Computing Surveys},
  volume  = {50},
  number  = {2},
  pages   = {21:1--21:35},
  year    = {2017},
  doi     = {10.1145/3054912}
}

@article{osa2018algorithmic,
  title   = {An Algorithmic Perspective on Imitation Learning},
  author  = {Osa, Takayuki and Pajarinen, Joni and Neumann, Gerhard and Bagnell, J. Andrew and Abbeel, Pieter and Peters, Jan},
  journal = {Foundations and Trends in Robotics},
  volume  = {7},
  number  = {1--2},
  pages   = {1--179},
  year    = {2018},
  doi     = {10.1561/2300000053}
}

@inproceedings{ho2016generative,
  title     = {Generative Adversarial Imitation Learning},
  author    = {Ho, Jonathan and Ermon, Stefano},
  booktitle = {Advances in Neural Information Processing Systems},
  volume    = {29},
  pages     = {4565--4573},
  year      = {2016},
  url       = {https://papers.nips.cc/paper/2016/hash/cc7e2b878868cbae992d1fb743995d8f-Abstract.html}
}

@inproceedings{kostrikov2020imitation,
  title     = {Imitation Learning via Off-Policy Distribution Matching},
  author    = {Kostrikov, Ilya and Nachum, Ofir and Tompson, Jonathan},
  booktitle = {International Conference on Learning Representations},
  year      = {2020},
  url       = {https://iclr.github.io/build/virtual/poster_Hyg-JC4FDr.html}
}

@inproceedings{kumar2022should,
  title     = {Should I Run Offline Reinforcement Learning or Behavioral Cloning?},
  author    = {Kumar, Aviral and Hong, Joey and Singh, Anikait and Levine, Sergey},
  booktitle = {International Conference on Learning Representations},
  year      = {2022},
  url       = {https://openreview.net/forum?id=AP1MKT37rJ}
}

@inproceedings{xu2022discriminator,
  title     = {Discriminator-Weighted Offline Imitation Learning from Suboptimal Demonstrations},
  author    = {Xu, Haoran and Zhan, Xianyuan and Yin, Honglei and Qin, Huiling},
  booktitle = {Proceedings of the 39th International Conference on Machine Learning},
  series    = {Proceedings of Machine Learning Research},
  volume    = {162},
  pages     = {24725--24742},
  year      = {2022},
  publisher = {PMLR},
  url       = {https://proceedings.mlr.press/v162/xu22l.html}
}

@book{sugiyama2012density,
  title     = {Density Ratio Estimation in Machine Learning},
  author    = {Sugiyama, Masashi and Suzuki, Taiji and Kanamori, Takafumi},
  publisher = {Cambridge University Press},
  year      = {2012},
  doi       = {10.1017/CBO9781139035613}
}

@article{kanamori2009ulsif,
  title   = {A Least-Squares Approach to Direct Importance Estimation},
  author  = {Kanamori, Takafumi and Hido, Shohei and Sugiyama, Masashi},
  journal = {Journal of Machine Learning Research},
  volume  = {10},
  number  = {48},
  pages   = {1391--1445},
  year    = {2009},
  url     = {https://jmlr.org/papers/v10/kanamori09a.html}
}

@inproceedings{nachum2019dualdice,
  title     = {{DualDICE}: Behavior-Agnostic Estimation of Discounted Stationary Distribution Corrections},
  author    = {Nachum, Ofir and Chow, Yinlam and Dai, Bo and Li, Lihong},
  booktitle = {Advances in Neural Information Processing Systems},
  volume    = {32},
  year      = {2019},
  url       = {https://proceedings.neurips.cc/paper/2019/hash/cf9a242b70f45317ffd281241fa66502-Abstract.html}
}

@article{nachum2019algaedice,
  title   = {{AlgaeDICE}: Policy Gradient from Arbitrary Experience},
  author  = {Nachum, Ofir and Dai, Bo and Kostrikov, Ilya and Chow, Yinlam and Li, Lihong and Schuurmans, Dale},
  journal = {arXiv preprint arXiv:1912.02074},
  year    = {2019},
  url     = {https://arxiv.org/abs/1912.02074}
}

@inproceedings{zhang2020gendice,
  title     = {{GenDICE}: Generalized Offline Estimation of Stationary Values},
  author    = {Zhang, Ruiyi and Dai, Bo and Li, Lihong and Schuurmans, Dale},
  booktitle = {International Conference on Learning Representations},
  year      = {2020},
  url       = {https://openreview.net/forum?id=HkxlcnVFwB}
}

@inproceedings{eysenbach2022contrastive,
  title     = {Contrastive Learning as Goal-Conditioned Reinforcement Learning},
  author    = {Eysenbach, Benjamin and Zhang, Tianjun and Salakhutdinov, Ruslan and Levine, Sergey},
  booktitle = {Advances in Neural Information Processing Systems},
  volume    = {35},
  year      = {2022},
  url       = {https://papers.nips.cc/paper_files/paper/2022/hash/e7663e974c4ee7a2b475a4775201ce1f-Abstract-Conference.html}
}

@inproceedings{zhang2020robust,
  title     = {Robust Deep Reinforcement Learning against Adversarial Perturbations on State Observations},
  author    = {Zhang, Huan and Chen, Hongge and Xiao, Chaowei and Li, Bo and Liu, Mingyan and Boning, Duane and Hsieh, Cho-Jui},
  booktitle = {Advances in Neural Information Processing Systems},
  volume    = {33},
  pages     = {21024--21037},
  year      = {2020},
  url       = {https://proceedings.neurips.cc/paper/2020/hash/f0eb6568ea114ba6e293f903c34d7488-Abstract.html}
}

\end{document}